\documentclass[sigconf]{acmart}
\AtBeginDocument{%
  \providecommand\BibTeX{{%
    \normalfont B\kern-0.5em{\scshape i\kern-0.25em b}\kern-0.8em\TeX}}}

\copyrightyear{2023} 
\acmYear{2023} 
\setcopyright{acmlicensed}\acmConference[ICBDT 2023]{2023 6th International Conference on Big Data Technologies (ICBDT)}{September 22--24, 2023}{Qingdao, China}
\acmBooktitle{2023 6th International Conference on Big Data Technologies (ICBDT) (ICBDT 2023), September 22--24, 2023, Qingdao, China}
\acmPrice{15.00}
\acmDOI{10.1145/3627377.3627442}
\acmISBN{979-8-4007-0766-7/23/09}

\acmPrice{15.00}
\acmISBN{978-1-4503-XXXX-X/18/06}

\begin{document}

\title{Enhancing Multimodal Understanding with CLIP-Based Image-to-Text Transformation}


\author{Chang Che}
\email{liamche1123@outlook.com}
\affiliation{%
  \institution{The George Washington University}
  \city{Atlanta}
  \state{Georgia}
  \country{USA}
}

\author{Qunwei Lin}
\email{linqunwei1030@outlook.com}
\affiliation{%
  \institution{Trine University}
  \city{Phoenix}
  \state{Arizona}
  \country{USA}
}

\author{Xinyu Zhao}
\email{lution798@gmail.com}
\affiliation{%
  \institution{Trine University}
  \city{Phoenix}
  \state{Arizona}
  \country{USA}
}

\author{Jiaxin Huang}
\email{jiaxinhuang1013@gmail.com}
\affiliation{%
 \institution{Trine University}
  \city{Phoenix}
  \state{Arizona}
  \country{USA}
}

\author{Liqiang Yu}
\email{rexyu@outlook.com}
\affiliation{%
  \institution{The University of Chicago}
  \city{Irvine}
  \state{California}
  \country{USA}}

\renewcommand{\shortauthors}{Chang and Qunwei, et al.}

\begin{abstract}
  The process of transforming input images into corresponding textual explanations stands as a crucial and complex endeavor within the domains of computer vision and natural language processing. In this paper, we propose an innovative ensemble approach that harnesses the capabilities of Contrastive Language-Image Pretraining (CLIP) models. Our ensemble framework encompasses two significant variations of the CLIP model, each meticulously designed to cater to specific nuances within the image-to-text transformation landscape. The first model introduces an elaborated architecture, featuring multiple layers with distinct learning rates, thereby amplifying its adeptness in capturing intricate relationships between images and text. The second model strategically exploits CLIP's inherent zero-shot learning potential to generate image-text embeddings, subsequently harnessed by a K-Nearest Neighbors (KNN) model. Through this KNN-based paradigm, the model facilitates image-to-text transformation by identifying closely related embeddings within the embedding space. Notably, our ensemble approach is rigorously evaluated, employing the cosine similarity metric to gauge the alignment between model-generated embeddings and ground truth representations. Comparative experiments vividly highlight the superiority of our ensemble strategy over standalone CLIP models. This study not only advances the state-of-the-art in image-to-text transformation but also accentuates the promising trajectory of ensemble learning in effectively addressing intricate multimodal tasks.
\end{abstract}





\maketitle

\section{Introduction}
    In the dynamic landscape of CV(computer vision) and NLP(natural language processing), the conversion of visual input into coherent textual descriptions has emerged as a fundamental yet intricate challenge. Bridging the gap between these two modalities holds profound potential across diverse domains, from enabling visually impaired individuals to enhancing the autonomy of machines. This article addresses the complex task of image-to-text transformation by introducing a novel ensemble approach that harnesses the capabilities of Contrastive Language-Image Pretraining (CLIP) models.
    
    While CLIP models have demonstrated exceptional prowess in aligning text and images, this work propels the field forward by presenting a sophisticated ensemble strategy that leverages their collective strengths. We introduce two distinct adaptations of the CLIP model, meticulously tailored to different facets of the image-to-text transformation endeavor. The first adaptation introduces a multi-layered architecture with the integration of differential learning rates, amplifying the model's discerning power to capture intricate and nuanced image-text relationships. Complementing this, the second adaptation ingeniously exploits CLIP's inherent zero-shot learning capacity, resulting in the generation of contextual embeddings for both images and text. These embeddings seamlessly integrate into a K-Nearest Neighbors (KNN)\cite{peterson2009k} model, a strategic choice that reverberates throughout the image-to-text transformation process.
    
    The significance of this study reverberates across numerous real-world scenarios, ranging from generating image captions to facilitating content-based image retrieval. Our ensemble approach, by bridging the semantic gap between images and text, contributes to the interpretability and accessibility of visual data.
    
    In subsequent sections, we embark on an exploratory journey into the intricate mechanics of our ensemble approach. We delve into architectural intricacies, elucidate data preprocessing methodologies, unveil the metrics employed for evaluation, and present compelling experimental results. Through meticulous evaluation and comprehensive comparative analysis, we substantiate the distinct advantages of our approach over both standalone CLIP models and traditional methods. As we navigate the complex landscape of image-to-text transformation, our study not only advances the boundaries of current capabilities but also underscores the burgeoning potential of ensemble learning to unravel the complexities inherent in multimodal tasks.

\section{Related Work}
    Within the domain of converting images to text, extensive research has been undertaken to improve the caliber and pertinence of the generated captions. This section provides an overview of key advancements and approaches that have contributed to the development of our proposed ensemble model for image captioning.
  
    J Devlin et al.\cite{devlin2018bert} introduce BERT, a transformer-based language model pre-trained on extensive text data. BERT's bidirectional context comprehension and contextual embeddings have propelled it to achieve state-of-the-art performance across diverse NLP tasks, including image captioning, significantly enhancing language understanding for such tasks.
    A Karpathy et al.\cite{karpathy2015deep} propose an attention-based model that aligns visual and semantic spaces, producing detailed image descriptions by focusing on relevant regions of the image. By allowing the model to attend to specific image regions while generating each word, the approach improves the relevance and contextual understanding of generated captions.
    Q. Wu et al. (2016) explore high-level concepts' role in bridging vision and language, specifically their impact on image captioning. By incorporating high-level semantic concepts, the model gains a better understanding of the image content, leading to improved caption quality that is more aligned with human perception.
    J Gu et al.\cite{gu2018stack}  propose a stack-captioning model that employs a coarse-to-fine approach to generate captions, demonstrating improved performance by iteratively refining the captioning process. Utilizing a hierarchy of captioning modules, this approach enables the model to encompass global and local details, yielding captions that are more contextually enriched and coherent.
    SJ Rennie et al.\cite{rennie2017self}   introduces self-critical sequence training, a reinforcement learning approach for image captioning, which improves caption quality by directly optimizing caption-level evaluation metrics. By using a reinforcement learning framework to fine-tune the captioning model, this approach encourages the generation of captions that receive higher scores according to the chosen evaluation metric.
    P Sharma et al.\cite{sharma2018conceptual} present the Conceptual Captions dataset, enhancing image caption quality and model training through improved annotations and hypernymic captions.
    J Johnson et al.\cite{johnson2016densecap} introduce DenseCap, a model merging convolutional and recurrent networks for dense image captioning. It generates captions for multiple image regions, enhancing description detail by covering various objects and regions within images.
    D Elliott et al.\cite{elliott2016multi30k} introduce the Multi30K dataset, a multilingual extension of the English-German image description dataset, facilitating cross-lingual research in image captioning. This dataset enables the evaluation and development of captioning models for multiple languages, contributing to the advancement of multilingual captioning.
    R Vedantam et al.\cite{vedantam2015cider}proposes the CIDEr metric, a consensus-based evaluation measure for image captioning that considers consensus among multiple reference captions, addressing limitations of previous metrics. CIDEr takes into account multiple valid caption variations and provides a more robust and comprehensive evaluation of caption quality.
    Utilizing the Twins-PCPVT model, Weinan Dai et al.\cite{dai2023diabetic} converts fundus images into embeddings, enhancing the efficiency and accuracy of diabetic retinopathy detection.
    A Radford et al.\cite{radford2015unsupervised} Generative Adversarial Networks (GANs) showcased potential in generating images from text descriptions. However, the unidirectional nature of the approach restricted its applicability to image-to-text tasks.
    Saad M.Darwith et al.\cite{darwish2015observations} investigate the use of Type-2 Fuzzy Logic as a means to bridge the semantic gap in Content-Based Image Retrieval (CBIR) systems. Their findings highlight the potential of this approach in enhancing image retrieval accuracy by addressing inherent uncertainties in semantic interpretations.
    J Lu et al.\cite{lu2019vilbert}showcased advancements in vision-and-language tasks by jointly pretraining on multiple vision and language tasks. However, the focus was broader, and the fine-grained image-to-text transformation remained a challenge.
    
    In the subsequent sections, we delve into our ensemble approach, unveiling its architectural intricacies, data preprocessing strategies, evaluation metrics, and empirical findings. Through comprehensive analysis, we highlight the distinct advantages of our approach, thereby contributing to the broader landscape of image-to-text transformation.
    
\section{Algorithm and Model}
    Our ensemble learning approach harnesses the power of two intricately designed models, both of which are built upon the foundation of the CLIP framework. Collectively, they collaborate to amplify the image-to-text transformation process, harnessing the exceptional abilities of CLIP.

\subsection{CLIP Model}
\begin{figure}[h]
        \centering
        \includegraphics[width=0.4\textwidth]{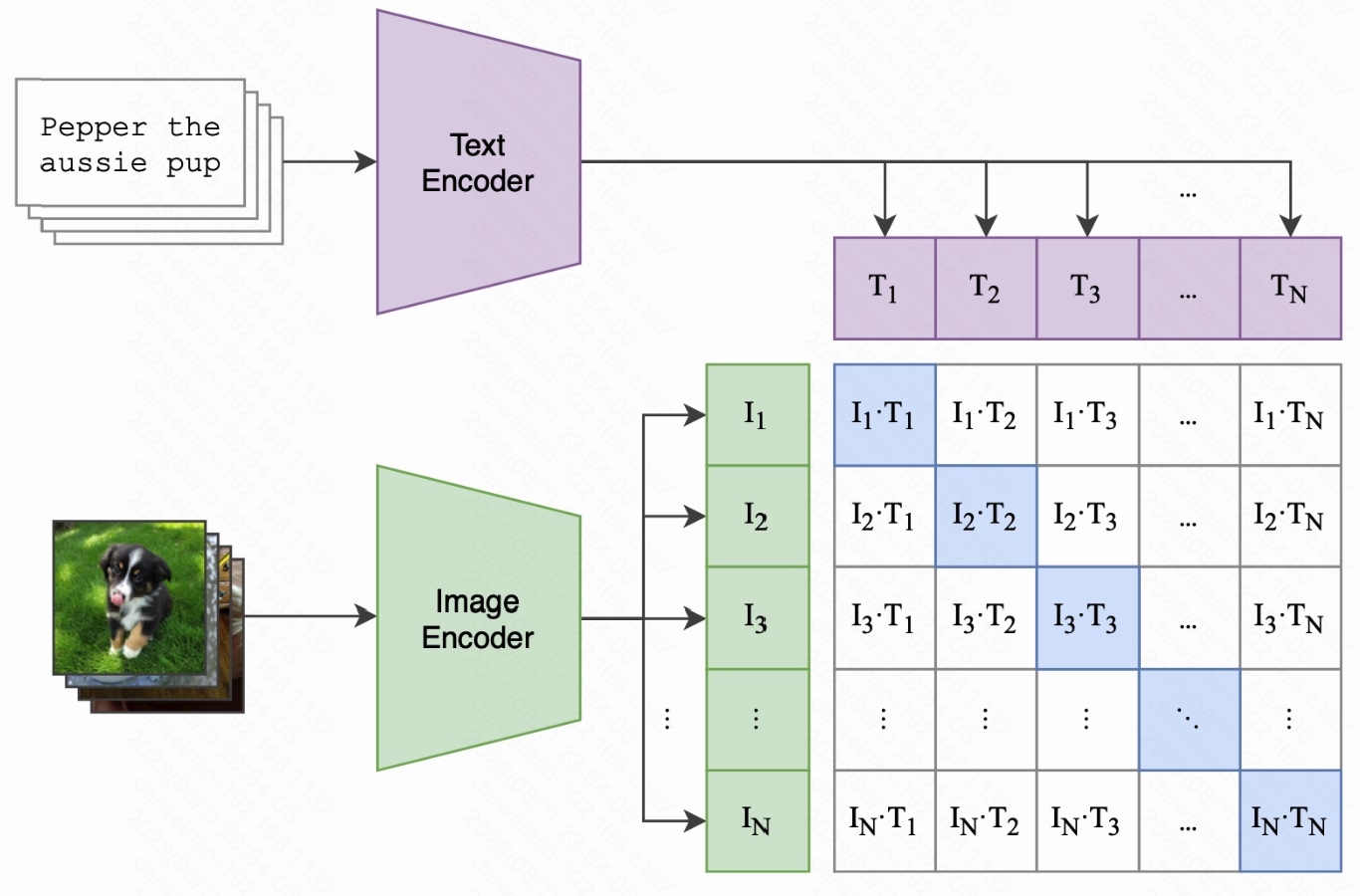}
        \caption{Architecture of the CLIP Model}
        \label{fig:clip_architecture}
    \end{figure}

    The CLIP model, which incorporates elements of the Vision Transformer (VIT), forms the foundation of our approach, providing a powerful framework for processing both images and text and generating embeddings in a shared semantic space. The architecture of the CLIP model, which includes VIT, is illustrated in Figure \ref{fig:clip_architecture}\cite{radford2021learning}.
    
    The CLIP model consists of a shared vision and language encoder. Leveraging the capabilities of VIT, the model employs self-attention mechanisms to capture intricate details within images. This integration allows CLIP to effectively process and represent visual information. Given an input image and text, the model projects them into a common space where the similarity between their embeddings reflects their semantic correspondence. This inherent ability of CLIP, enhanced by VIT, forms the basis for our subsequent enhancements.

    The Vision Transformer (VIT)\cite{yuan2021tokens} is a crucial component integrated into the CLIP model. VIT is an image processing model that utilizes self-attention mechanisms to analyze and capture the relationships between different parts of an image. By incorporating VIT, the CLIP model gains the capability to understand images at a more granular level, enabling it to extract meaningful visual features and representations. This integration not only enhances the model's ability to process images but also contributes to the generation of semantically meaningful embeddings.

\subsection{Model A - Enhanced Image-to-Text Transformation}
    To optimize the process of image-to-text transformation, we introduce Model A, which enhances the capabilities of the CLIP model through the incorporation of additional neural network layers. Model A is designed to refine the visual embedding of an input image and generate an enriched text embedding.
    
    The computation within Model A begins with passing the input image through the pre-trained CLIP vision model, resulting in a high-dimensional visual embedding denoted as \( \text{Image\_Embedding} \). This visual representation captures the salient features of the image in the context of semantic understanding.
    
    \begin{figure}[h]
        \centering
        \includegraphics[width=0.48\textwidth]{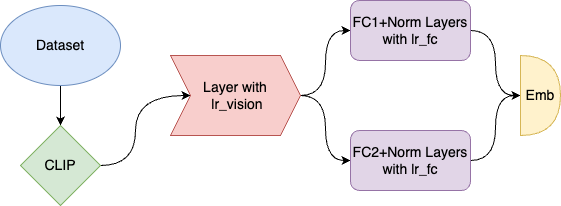}
        \caption{Architecture of Model A - Enhanced Image-to-Text Transformation}
        \label{fig:model_a_architecture}
    \end{figure}
    
    To enhance the visual embedding, Model A introduces two additional fully connected layers, \( \text{FC1} \) and \( \text{FC2} \), followed by layer normalization operations \( \text{Norm1} \) and \( \text{Norm2} \). These layers iteratively refine the feature representation, extracting intricate relationships inherent in the image data.
    
    The out of \( \text{FC1} \) is computed as follows:
    
    \begin{equation}
    \text{FC1\_Out} = \text{FC1}(\text{Image\_Emb})
    \end{equation}
    
    Subsequently, the output is normalized using layer normalization to yield \( \text{Norm1\_Out} \):
    
    \begin{equation}
    \text{Norm1\_Out} = \text{Norm1}(\text{FC1\_Out})
    \end{equation}
    
    Similarly, \( \text{FC2} \) and \( \text{Norm2} \) operations further enhance the features, resulting in \( \text{Norm2\_Out} \).
    
    To produce the enriched text emb, Model A employs a weighted combination of \( \text{Norm1\_Out} \) and \( \text{Norm2\_Out} \) using the following equation:
    
    \begin{equation}
    \text{Final\_Text\_Emb} = \alpha \cdot \text{Norm1\_Out} + (1 - \alpha) \cdot \text{Norm2\_Out}
    \end{equation}
    
    This fusion mechanism empowers Model A to adaptively integrate the refined visual features, resulting in an enriched text embedding that encapsulates both the original visual content and its semantic nuances.
    
    To facilitate effective training, we employ a differential learning rate strategy. Specifically, the pre-trained CLIP vision model is fine-tuned using a relatively small learning rate \( lr_{\text{vision}} \), while the newly introduced fully connected layers \( \text{FC1} \) and \( \text{FC2} \), along with layer normalization operations \( \text{Norm1} \) and \( \text{Norm2} \), use a larger learning rate \( lr_{\text{fc}} \). This method guarantees a harmonious parameter adjustment between the existing pre-trained model and the freshly incorporated layers. The pre-trained CLIP model, having been optimized on an extensive dataset, demands nuanced tweaks, while the newly added layers necessitate more substantial updates owing to their random initialization.

\subsection{Model B - Zero-Shot Learning and KNN-based Fusion}
    Model B capitalizes on the inherent zero-shot learning \cite{xian2017zero} capabilities of the CLIP model, extending its applicability to image-to-text transformation tasks. The architecture of Model B integrates image embeddings with their corresponding text embeddings through a K-nearest neighbors (KNN) based fusion approach. This enables a seamless cross-modal interaction, leveraging the rich semantic understanding of CLIP for enhanced image-to-text transformation.
    
    The first step of Model B\cite{radford2021learning} involves utilizing the CLIP model to generate both image embeddings and text embeddings for a diverse set of images and their corresponding textual descriptions. This forms a basis for zero-shot learning, enabling Model B to generalize to unseen image-text pairs during testing.
    
    \begin{figure}[h]
        \centering
        \includegraphics[width=0.4\textwidth]{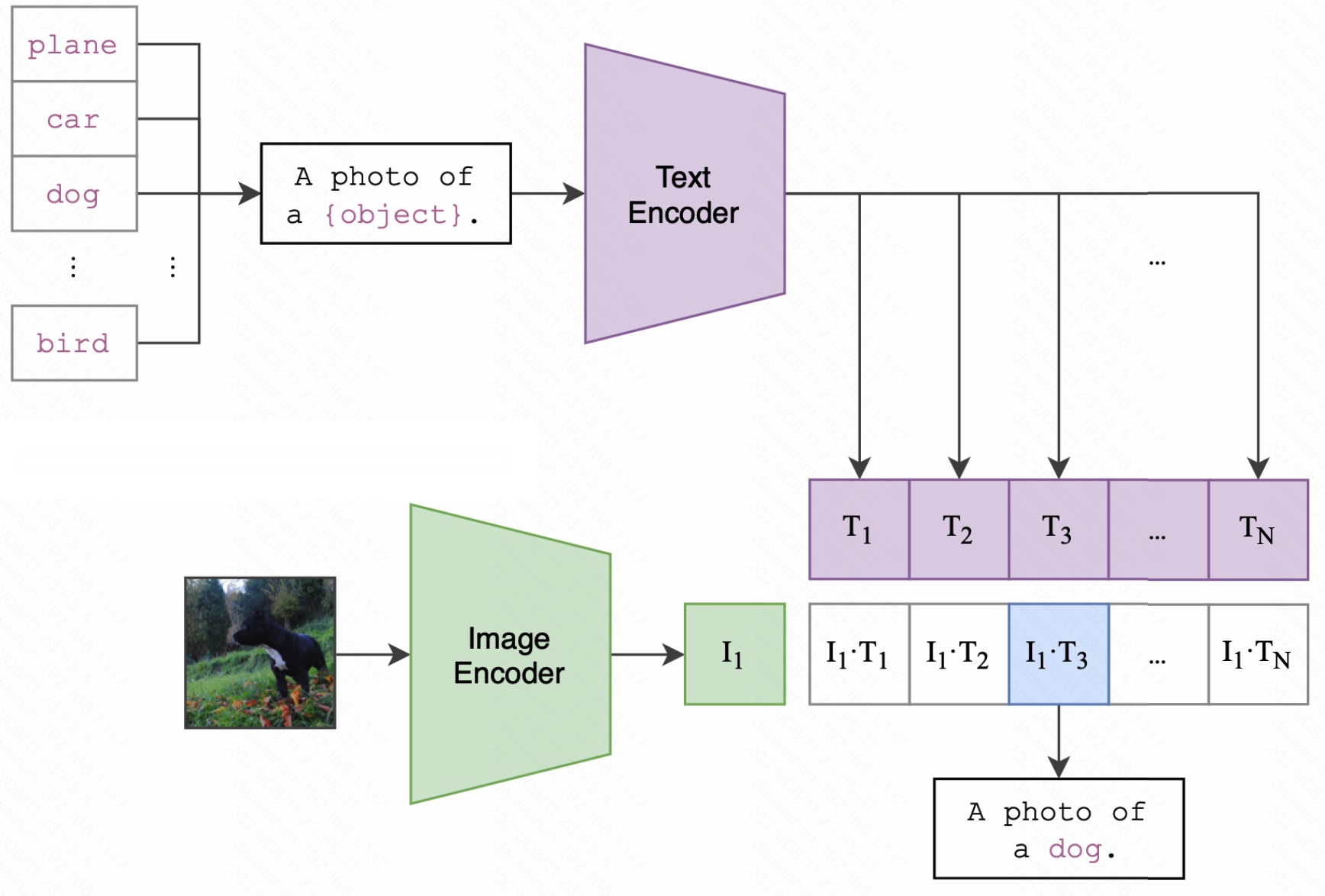}
        \caption{Zero-Shot Learning Architecture in Model B}
        \label{fig:zeroshot_structure}
    \end{figure}
    
    For each image embedding, Model B employs a K-nearest neighbors (KNN) model to retrieve a set of nearest neighbor text embeddings. The KNN-based approach introduces a distance-based weighting mechanism, which captures the relevance and contextual significance of each neighbor. This distance-based weighting is calculated using the following formula:
    
    \begin{equation}
    \text{Weight}(i) = \frac{1}{\text{Distance}(i)^{\text{Distance\_Dim}} + \delta} \times \text{Coef}
    \end{equation}
    
    Where \( \text{Distance}(i) \) is the Euclidean distance between the query image embedding and the \( i \)th neighbor text embedding, \( \text{Distance\_Dim} \) controls the influence of distance, and \( \delta \) is a small positive constant to ensure numerical stability, and \( \text{Coef} \) is a coefficient to prevent overflow.
    
    The weighted contributions from all neighbors are then aggregated to form the final text embedding. Mathematically, given an image embedding \( \text{Image\_Emb} \), the KNN-based fusion mechanism generates the text embedding \( \text{Text\_Emb} \) as:
    
    \begin{equation}
    \text{Text\_Emb} = \frac{1}{K} \sum_{i=1}^{K} \text{Weight}(i) \times \text{KNN\_Text\_Emb}_i
    \end{equation}
    
    Here, \( K \) represents the number of nearest neighbors, \( \text{KNN\_Text\_Emb}_i \) refers to the \( i \)th nearest neighbor text embedding.
    
    In summary, Model B extends the zero-shot learning capabilities of the CLIP model to image-to-text transformation. The KNN-based fusion approach intelligently combines image and text emb, with distance-based weighting to capture contextual relevance, resulting in an enriched text representation that reflects the underlying semantics.
    
    \begin{figure}[h]
        \centering
        \includegraphics[width=0.48\textwidth]{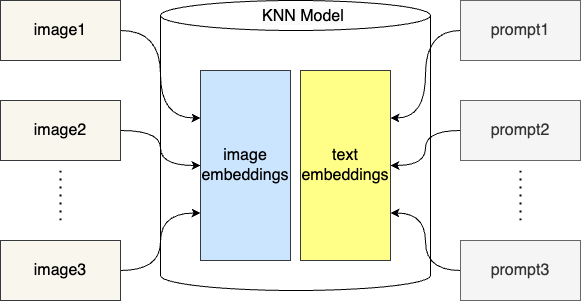}
        \caption{K-Nearest Neighbors Fusion in Model B}
        \label{fig:knn_structure}
    \end{figure}

\subsection{Model Ensemble}
    This ensemble leverages the strengths of each constituent, yielding a powerful image-to-text transformation framework.
    
    The integration process merges the semantically rich embeddings from the trained CLIP model with the contextual embeddings produced by the CLIP kNNRegression model. This fusion aims to leverage CLIP's interpretive capabilities alongside the nuanced contextual understanding of kNNRegression, creating a unified ensemble embedding.  The ensemble embedding is calculated as follows:
    
    \begin{equation}
        \text{Ens\_Emb} = \alpha \times \text{A\_Emb} + (1 - \alpha) \times \text{B\_Emb}
    \end{equation}
    
    Where \(\alpha\) represents the adjusted weighting coefficient.
    
    The model ensemble embodies the symbiotic relationship between the model A and model B, resulting in a dynamic and versatile image-to-text transformation framework. The subsequent sections delve into empirical evaluation and findings, substantiating the effectiveness and potential of our ensemble approach.

\subsection{Data Preprocessing}
    
    Effective data preprocessing is pivotal to the success of our ensemble learning approach for image-to-text transformation. We employ a series of meticulous steps to enhance the quality and suitability of our input data.
    
    \subsubsection{Data Augmentation}
    
        Data augmentation is employed to enhance the diversity and robustness of our training dataset. For images, we apply random transformations such as rotations, flips, and crops to generate augmented versions of the original images. For text, we perform synonym replacement and random word shuffling to create variations of the input text descriptions.
    
    \subsubsection{Cosine Similarity Filtering}
        To streamline the dataset and enhance model performance, we transform textual prompts into embedding. These embedding encapsulate intricate semantic information within the embedding space. To quantify the resemblance between two embedding vectors, we utilize the cosine similarity formula:
        
        \begin{equation}
        \text{Cosine Similarity}(v_i, v_j) = \frac{v_i \cdot v_j}{\|v_i\| \cdot \|v_j\|}
        \end{equation}
        
        Here, $v_i$ and $v_j$ denote the embedding vectors associated with two distinct textual prompts.
        
        For each textual prompt within the training dataset, we compute the cosine value with the embedding vectors of other prompts. By applying a designated similarity threshold (e.g., 0.85), in cases where the similarity between a prompt and any other prompt surpasses this threshold, we opt to remove one of the samples to ensure a diverse dataset composition.
        
        This filtering step not only enhances model performance but also leads to improved training efficiency. Specifically, the reduction in dataset size from 90K to 60K instances has resulted in an enhancement of the model's metric, along with a reduction in training time per epoch from 6 hours to 4 hours. These improvements collectively contribute to the effectiveness and efficiency of our ensemble learning approach for image-to-text transformation tasks.

\subsection{Performance Measurement}

    To gauge model performance, we utilize the following metric:
    
    \begin{equation}
    \text{Avg-Cos} = \frac{1}{N} \sum_{i=1}^{N} \text{CosSim}(\text{GT-Embed}_i, \text{Pred-Embed}_i)
    \end{equation}
    
    Where $\text{Avg-Cos}$ denotes the average cosine similarity, $N$ signifies the total image-text pairs, and $\text{CosSim}(\text{GT-Embed}_i, \text{Pred-Embed}_i)$ represents the cosine value between the groundtruth text embedding and the predicted embedding for the $i$-th pair. This metric quantifies the alignment and semantic coherence between images and their textual descriptions. Higher $\text{Avg-CosSim}$ values indicate more robust semantic alignment and enhanced image-to-text transformation, showcasing the effectiveness of our ensemble learning approach.

    \subsection{Experiment Results}
    
    We conducted comprehensive experiments to evaluate the performance of our proposed ensemble approach for image-to-text transformation. In our experiments, we employ the Average Cosine Similarity (Avg-CosSim) as the evaluation metric, assessing the correspondence between the embeddings generated by the model and the actual ground truth representations.

    \begin{table}
      \caption{Average Cosine Similarity Results}
      \label{tab:results}
      \begin{tabular}{cc}
        \toprule
        Model&Avg-CosSim\\
        \midrule
        CLIP & 0.5642 \\
        CLIP with data filter1 & 0.5684 \\
        CLIP with data filter2 & 0.5721 \\
        Model A & 0.5753 \\
        Model B & 0.5531 \\
        Ensemble Model & 0.5961 \\
      \bottomrule
    \end{tabular}
    \end{table}
    
    The outcomes presented in Table \ref{tab:results} unmistakably showcase the efficacy of our ensemble approach. The Ensemble Model achieves the highest Avg-CosSim, indicating a superior alignment between the generated embeddings and ground truth representations compared to other individual models. Model A, with its elaborated architecture, also outperforms the standalone CLIP models and Model B. These findings underscore the value of our ensemble strategy in enhancing image-to-text transformation.
    
    The comparative experiments validate the promising trajectory of ensemble learning in addressing intricate multimodal tasks and further contribute to advancing the state-of-the-art in image-to-text transformation.

\bibliographystyle{ACM-Reference-Format}
\bibliography{sample-base}


\begin{thebibliography}{16}


\ifx \showCODEN    \undefined \def \showCODEN     #1{\unskip}     \fi
\ifx \showDOI      \undefined \def \showDOI       #1{#1}\fi
\ifx \showISBNx    \undefined \def \showISBNx     #1{\unskip}     \fi
\ifx \showISBNxiii \undefined \def \showISBNxiii  #1{\unskip}     \fi
\ifx \showISSN     \undefined \def \showISSN      #1{\unskip}     \fi
\ifx \showLCCN     \undefined \def \showLCCN      #1{\unskip}     \fi
\ifx \shownote     \undefined \def \shownote      #1{#1}          \fi
\ifx \showarticletitle \undefined \def \showarticletitle #1{#1}   \fi
\ifx \showURL      \undefined \def \showURL       {\relax}        \fi
\providecommand\bibfield[2]{#2}
\providecommand\bibinfo[2]{#2}
\providecommand\natexlab[1]{#1}
\providecommand\showeprint[2][]{arXiv:#2}

\bibitem[Dai et~al\mbox{.}(2023)]%
        {dai2023diabetic}
\bibfield{author}{\bibinfo{person}{Weinan Dai}, \bibinfo{person}{Chengjie Mou}, \bibinfo{person}{Jun Wu}, {and} \bibinfo{person}{Xuesong Ye}.} \bibinfo{year}{2023}\natexlab{}.
\newblock \showarticletitle{Diabetic Retinopathy Detection with Enhanced Vision Transformers: The Twins-PCPVT Solution}. In \bibinfo{booktitle}{\emph{2023 IEEE 3rd International Conference on Electronic Technology, Communication and Information (ICETCI)}}. IEEE, \bibinfo{pages}{403--407}.
\newblock


\bibitem[Darwish and Ali(2015)]%
        {darwish2015observations}
\bibfield{author}{\bibinfo{person}{Saad~M Darwish} {and} \bibinfo{person}{Raad~A Ali}.} \bibinfo{year}{2015}\natexlab{}.
\newblock \showarticletitle{Observations on using type-2 fuzzy logic for reducing semantic gap in content--based image retrieval system}.
\newblock \bibinfo{journal}{\emph{International Journal of Computer Theory and Engineering}} \bibinfo{volume}{7}, \bibinfo{number}{1} (\bibinfo{year}{2015}), \bibinfo{pages}{1}.
\newblock


\bibitem[Devlin et~al\mbox{.}(2018)]%
        {devlin2018bert}
\bibfield{author}{\bibinfo{person}{Jacob Devlin}, \bibinfo{person}{Ming-Wei Chang}, \bibinfo{person}{Kenton Lee}, {and} \bibinfo{person}{Kristina Toutanova}.} \bibinfo{year}{2018}\natexlab{}.
\newblock \showarticletitle{Bert: Pre-training of deep bidirectional transformers for language understanding}.
\newblock \bibinfo{journal}{\emph{arXiv preprint arXiv:1810.04805}} (\bibinfo{year}{2018}).
\newblock


\bibitem[Elliott et~al\mbox{.}(2016)]%
        {elliott2016multi30k}
\bibfield{author}{\bibinfo{person}{Desmond Elliott}, \bibinfo{person}{Stella Frank}, \bibinfo{person}{Khalil Sima'an}, {and} \bibinfo{person}{Lucia Specia}.} \bibinfo{year}{2016}\natexlab{}.
\newblock \showarticletitle{Multi30k: Multilingual english-german image descriptions}.
\newblock \bibinfo{journal}{\emph{arXiv preprint arXiv:1605.00459}} (\bibinfo{year}{2016}).
\newblock


\bibitem[Gu et~al\mbox{.}(2018)]%
        {gu2018stack}
\bibfield{author}{\bibinfo{person}{Jiuxiang Gu}, \bibinfo{person}{Jianfei Cai}, \bibinfo{person}{Gang Wang}, {and} \bibinfo{person}{Tsuhan Chen}.} \bibinfo{year}{2018}\natexlab{}.
\newblock \showarticletitle{Stack-captioning: Coarse-to-fine learning for image captioning}. In \bibinfo{booktitle}{\emph{Proceedings of the AAAI conference on artificial intelligence}}, Vol.~\bibinfo{volume}{32}.
\newblock


\bibitem[Johnson et~al\mbox{.}(2016)]%
        {johnson2016densecap}
\bibfield{author}{\bibinfo{person}{Justin Johnson}, \bibinfo{person}{Andrej Karpathy}, {and} \bibinfo{person}{Li Fei-Fei}.} \bibinfo{year}{2016}\natexlab{}.
\newblock \showarticletitle{Densecap: Fully convolutional localization networks for dense captioning}. In \bibinfo{booktitle}{\emph{Proceedings of the IEEE conference on computer vision and pattern recognition}}. \bibinfo{pages}{4565--4574}.
\newblock


\bibitem[Karpathy and Fei-Fei(2015)]%
        {karpathy2015deep}
\bibfield{author}{\bibinfo{person}{Andrej Karpathy} {and} \bibinfo{person}{Li Fei-Fei}.} \bibinfo{year}{2015}\natexlab{}.
\newblock \showarticletitle{Deep visual-semantic alignments for generating image descriptions}. In \bibinfo{booktitle}{\emph{Proceedings of the IEEE conference on computer vision and pattern recognition}}. \bibinfo{pages}{3128--3137}.
\newblock


\bibitem[Lu et~al\mbox{.}(2019)]%
        {lu2019vilbert}
\bibfield{author}{\bibinfo{person}{Jiasen Lu}, \bibinfo{person}{Dhruv Batra}, \bibinfo{person}{Devi Parikh}, {and} \bibinfo{person}{Stefan Lee}.} \bibinfo{year}{2019}\natexlab{}.
\newblock \showarticletitle{Vilbert: Pretraining task-agnostic visiolinguistic representations for vision-and-language tasks}.
\newblock \bibinfo{journal}{\emph{Advances in neural information processing systems}}  \bibinfo{volume}{32} (\bibinfo{year}{2019}).
\newblock


\bibitem[Peterson(2009)]%
        {peterson2009k}
\bibfield{author}{\bibinfo{person}{Leif~E Peterson}.} \bibinfo{year}{2009}\natexlab{}.
\newblock \showarticletitle{K-nearest neighbor}.
\newblock \bibinfo{journal}{\emph{Scholarpedia}} \bibinfo{volume}{4}, \bibinfo{number}{2} (\bibinfo{year}{2009}), \bibinfo{pages}{1883}.
\newblock


\bibitem[Radford et~al\mbox{.}(2021)]%
        {radford2021learning}
\bibfield{author}{\bibinfo{person}{Alec Radford}, \bibinfo{person}{Jong~Wook Kim}, \bibinfo{person}{Chris Hallacy}, \bibinfo{person}{Aditya Ramesh}, \bibinfo{person}{Gabriel Goh}, \bibinfo{person}{Sandhini Agarwal}, \bibinfo{person}{Girish Sastry}, \bibinfo{person}{Amanda Askell}, \bibinfo{person}{Pamela Mishkin}, \bibinfo{person}{Jack Clark}, {et~al\mbox{.}}} \bibinfo{year}{2021}\natexlab{}.
\newblock \showarticletitle{Learning transferable visual models from natural language supervision}. In \bibinfo{booktitle}{\emph{International conference on machine learning}}. PMLR, \bibinfo{pages}{8748--8763}.
\newblock


\bibitem[Radford et~al\mbox{.}(2015)]%
        {radford2015unsupervised}
\bibfield{author}{\bibinfo{person}{Alec Radford}, \bibinfo{person}{Luke Metz}, {and} \bibinfo{person}{Soumith Chintala}.} \bibinfo{year}{2015}\natexlab{}.
\newblock \showarticletitle{Unsupervised representation learning with deep convolutional generative adversarial networks}.
\newblock \bibinfo{journal}{\emph{arXiv preprint arXiv:1511.06434}} (\bibinfo{year}{2015}).
\newblock


\bibitem[Rennie et~al\mbox{.}(2017)]%
        {rennie2017self}
\bibfield{author}{\bibinfo{person}{Steven~J Rennie}, \bibinfo{person}{Etienne Marcheret}, \bibinfo{person}{Youssef Mroueh}, \bibinfo{person}{Jerret Ross}, {and} \bibinfo{person}{Vaibhava Goel}.} \bibinfo{year}{2017}\natexlab{}.
\newblock \showarticletitle{Self-critical sequence training for image captioning}. In \bibinfo{booktitle}{\emph{Proceedings of the IEEE conference on computer vision and pattern recognition}}. \bibinfo{pages}{7008--7024}.
\newblock


\bibitem[Sharma et~al\mbox{.}(2018)]%
        {sharma2018conceptual}
\bibfield{author}{\bibinfo{person}{Piyush Sharma}, \bibinfo{person}{Nan Ding}, \bibinfo{person}{Sebastian Goodman}, {and} \bibinfo{person}{Radu Soricut}.} \bibinfo{year}{2018}\natexlab{}.
\newblock \showarticletitle{Conceptual captions: A cleaned, hypernymed, image alt-text dataset for automatic image captioning}. In \bibinfo{booktitle}{\emph{Proceedings of the 56th Annual Meeting of the Association for Computational Linguistics (Volume 1: Long Papers)}}. \bibinfo{pages}{2556--2565}.
\newblock


\bibitem[Vedantam et~al\mbox{.}(2015)]%
        {vedantam2015cider}
\bibfield{author}{\bibinfo{person}{Ramakrishna Vedantam}, \bibinfo{person}{C Lawrence~Zitnick}, {and} \bibinfo{person}{Devi Parikh}.} \bibinfo{year}{2015}\natexlab{}.
\newblock \showarticletitle{Cider: Consensus-based image description evaluation}. In \bibinfo{booktitle}{\emph{Proceedings of the IEEE conference on computer vision and pattern recognition}}. \bibinfo{pages}{4566--4575}.
\newblock


\bibitem[Xian et~al\mbox{.}(2017)]%
        {xian2017zero}
\bibfield{author}{\bibinfo{person}{Yongqin Xian}, \bibinfo{person}{Bernt Schiele}, {and} \bibinfo{person}{Zeynep Akata}.} \bibinfo{year}{2017}\natexlab{}.
\newblock \showarticletitle{Zero-shot learning-the good, the bad and the ugly}. In \bibinfo{booktitle}{\emph{Proceedings of the IEEE conference on computer vision and pattern recognition}}. \bibinfo{pages}{4582--4591}.
\newblock


\bibitem[Yuan et~al\mbox{.}(2021)]%
        {yuan2021tokens}
\bibfield{author}{\bibinfo{person}{Li Yuan}, \bibinfo{person}{Yunpeng Chen}, \bibinfo{person}{Tao Wang}, \bibinfo{person}{Weihao Yu}, \bibinfo{person}{Yujun Shi}, \bibinfo{person}{Zi-Hang Jiang}, \bibinfo{person}{Francis~EH Tay}, \bibinfo{person}{Jiashi Feng}, {and} \bibinfo{person}{Shuicheng Yan}.} \bibinfo{year}{2021}\natexlab{}.
\newblock \showarticletitle{Tokens-to-token vit: Training vision transformers from scratch on imagenet}. In \bibinfo{booktitle}{\emph{Proceedings of the IEEE/CVF international conference on computer vision}}. \bibinfo{pages}{558--567}.
\newblock


\end{thebibliography}

\appendix

\end{document}